\documentclass{article}

\usepackage[nonatbib,final]{neurips_2024}

\usepackage[utf8]{inputenc} 
\usepackage[T1]{fontenc}    
\usepackage{hyperref}       
\usepackage{url}            
\usepackage{booktabs}       
\usepackage{amsfonts}       
\usepackage{nicefrac}       
\usepackage{microtype}      
\usepackage{xcolor}         
\usepackage{graphicx}
\usepackage{amsmath}
\DeclareMathOperator*{\argmax}{argmax}
\usepackage{color, colortbl}
\usepackage{caption}
\usepackage{makecell}

\usepackage{algorithm}
\usepackage{algorithmic}
\newcommand\CoAuthorMark{\footnotemark[\arabic{footnote}]}
\usepackage{float}

\title{TALoS: Enhancing Semantic Scene Completion via Test-time Adaptation on the Line of Sight}

\author{%
  Hyun-Kurl Jang\thanks{denotes equal contribution.}\\
  Visual Intelligence Lab.\\
  KAIST\\
  \texttt{jhg0001@kaist.ac.kr} \\
  \And
  Jihun Kim\CoAuthorMark\\
  Visual Intelligence Lab.\\
  KAIST\\
  \texttt{jihun1998@kaist.ac.kr} \\
  \And
  Hyeokjun Kweon\CoAuthorMark\\
  Visual Intelligence Lab.\\
  KAIST\\
  \texttt{0327june@kaist.ac.kr} \\
  \And
  Kuk-Jin Yoon\\
  Visual Intelligence Lab.\\
  KAIST\\
  \texttt{kjyoon@kaist.ac.kr} \\
}

\begin{document}

\maketitle

\begin{abstract}
Semantic Scene Completion (SSC) aims to perform geometric completion and semantic segmentation simultaneously. 
Despite the promising results achieved by existing studies, the inherently ill-posed nature of the task presents significant challenges in diverse driving scenarios. 
This paper introduces TALoS, a novel test-time adaptation approach for SSC that excavates the information available in driving environments.
Specifically, we focus on that observations made at a certain moment can serve as Ground Truth (GT) for scene completion at another moment. 
Given the characteristics of the LiDAR sensor, an observation of an object at a certain location confirms both 1) the occupation of that location and 2) the absence of obstacles along the line of sight from the LiDAR to that point.
TALoS utilizes these observations to obtain self-supervision about occupancy and emptiness, guiding the model to adapt to the scene in test time. 
In a similar manner, we aggregate reliable SSC predictions among multiple moments and leverage them as semantic pseudo-GT for adaptation.
Further, to leverage future observations that are not accessible at the current time, we present a dual optimization scheme using the model in which the update is delayed until the future observation is available.
Evaluations on the SemanticKITTI validation and test sets demonstrate that TALoS significantly improves the performance of the pre-trained SSC model. Our code is available at \url{https://github.com/blue-531/TALoS}.
\end{abstract}

\section{Introduction}\label{Sec:intro}

LiDAR is a predominant 3D sensor in autonomous vehicles, effectively capturing the 3D geometry of the surroundings as a point cloud.
However, LiDAR inherently records the surface of objects, leaving the areas behind initial contact points empty.
Therefore, it is crucial for safety and driving planning to predict the state of these hidden regions using only the limited information available. 
Addressing these challenges, Semantic Scene Completion (SSC) has emerged as a pivotal research topic, enabling simultaneous geometric completion and semantic segmentation of the surroundings.

Existing SSC studies have focused on tackling both tasks~\cite{song2017semantic,rist2021semantic,xia2023scpnet,roldao2020lmscnet,yan2021sparse,yang2021semantic,zhang2018efficient,zou2021up,garbade2019two}, mainly from an architectural perspective. 
By amalgamating the models specialized for each task, these approaches have shown promising results over the last few years. 
Nevertheless, the nature of the completion task—filling in the unseen parts from the given observation—heavily relies on the prior structural distribution learned from the training dataset. 
Therefore, in our view, the classical SSC paradigm is inevitably vulnerable to handling the diverse scene structures encountered in driving scenarios.

As a remedy, this paper pioneers a novel SSC approach based on Test-time Adaptation (TTA), which adjusts a pre-trained model to adapt to each test environment.
Due to the absence of ground truths (GTs) during test time, the existing TTA studies for various fields have endeavored to design optimization goals, like meta-learning or auxiliary tasks~\cite{sun2020test,liu2021ttt++,huang2021model, mirza2023mate,hatem2023point}.
Instead, we focus on the driving scenarios assumed by SSC, excavating the information helpful for adapting the model to the scene in test time.

Our main idea is simple yet effective: \textbf{an observation made at one moment could serve as supervision for the SSC prediction at another moment}.
While traveling through an environment, an autonomous vehicle can continuously observe the overall scene structures, including objects that were previously occluded (or will be occluded later), which are concrete guidances for the adaptation of scene completion.
Given the characteristics of the LiDAR sensor, an observation of a point at a specific spatial location at a specific moment confirms not only the occupation at that location itself but also the absence of obstacles along the line of sight from the sensor to that location.

The proposed method, named \textbf{Test-time Adaptation via Line of Sight (TALoS)}, is designed to explicitly leverage these characteristics, obtaining self-supervision for geometric completion.
Additionally, we extend the TALoS framework for semantic recognition, another key goal of SSC, by collecting the reliable regions only among the semantic segmentation results predicted at each moment.
Further, to leverage valuable future information that is not accessible at the time of the current update, we devise a novel dual optimization scheme involving the model gradually updating across the temporal dimension.
This enables the model to continuously adapt to the surroundings at test time without any manual guidance, ultimately achieving better SSC performance.

We verify the superiority of TALoS on the SemanticKITTI~\cite{behley2019semantickitti} benchmark. 
The results strongly confirm that TALoS enhances not only geometric completion but also semantic segmentation performance by large margins.
With extensive experiments, including ablation studies, we analyze the working logic of TALoS in detail and present its potential as a viable solution for practical SSC. 


\section{Related Works}\label{Sec:rw}

\subsection{Semantic scene completion}

Starting from SSCNet~\cite{song2017semantic}, semantic scene completion task has been extensively studied~\cite{cheng2021s3cnet,rist2021semantic,xia2023scpnet,roldao2020lmscnet,yan2021sparse,yang2021semantic,zhang2018efficient,zou2021up,garbade2019two, li2023sscbench, tian2024occ3d, cao2022monoscene, shi2024panossc, li2023voxformer, zhang2023occformer, huang2023tri}. 
In SSC using LiDAR, as both tasks of geometric completion and semantic understanding should be achieved simultaneously, the existing studies have mainly presented architectural approaches.
For example, LMSC~\cite{roldao2020lmscnet} and UtD~\cite{zou2021up} utilize UNet-based structures with multi-scale connections. 
JS3C-Net~\cite{yan2021sparse} and SSA-SC~\cite{yang2021semantic} propose architectures consisting of semantic segmentation and completion networks to utilize them complementarily. 
Although these approaches show promise, handling the diversities inherent in outdoor scenes remains a challenging problem. 
In this light, we would like to introduce test-time adaptation to the field of semantic scene completion.

Notably, SCPNet~\cite{xia2023scpnet} proposes to use distillation during the training phase, transferring the knowledge from the model using multiple scans to the model using a single scan.
Although this approach also aims to use information from various moments, the proposed TALoS is distinct as it leverages such information online, adapting the model to the diverse driving sequence.
Also, the recently published OccFiner~\cite{shi2024occfiner} is noteworthy, as it aims to enhance the already existing SSC model.
However, OccFiner is a post-processing method that refines the results of the pre-trained model, performing in an offline manner, unlike our online TTA-based approach.

\subsection{Test-time adaptation}

Test-time Adaptation~(TTA) aims to adapt a pre-trained model to target data in test time, without access to the source domain data used for training. 
One widely used method involves attaching additional self-supervised learning branches to the model~\cite{sun2020test,liu2021ttt++,huang2021model}.
In point cloud TTAs, using auxiliary tasks such as point cloud reconstruction~\cite{mirza2023mate,hatem2023point} are actively studied. 
However, these approaches require the model to be trained with the additional branches, primarily in the training stage on the source dataset.

To relieve the requirements on the training stage, various online optimization goals have been explored, such as information maximization~\cite{wang2020tent,liang2020we,niu2023towards} and pseudo labeling~\cite{liang2020we,wang2022continual,fleuret2021uncertainty,tomar2023tesla} schemes.
Similar approaches have also been proposed for the point cloud, as in ~\cite{shin2022mm, cao2023multi}, using pseudo labeling.

Unfortunately, despite the natural fit between these TTA approaches and the goal of SSC, which involves completing diverse driving environments, the use of TTA has been scarcely explored in the SSC field.
Against this background, we pioneer the TTA-based SSC method, especially focusing on excavating the information from the point clouds consecutively observed at various moments.

\section{Method}\label{Sec:method}

\subsection{Problem definition}\label{Sec:pre}
This section begins by defining the formulation of our approach and the notations used throughout the paper. 
The Semantic Scene Completion (SSC) task aims to learn a mapping function from an input point cloud to the completed voxel representation. 
We formally denote the input point cloud $\mathbf{X}\in\mathbb{R}^{N\times3}$ as a set of points, where each point represents its XYZ coordinate. 
Following the conventional SSC studies, the completion result is denoted as $\mathbf{Y}\in \mathbb{C}^{L\times W \times H}$. 
Here, $L, W, H$ are the dimensions of the voxel grid, and $\mathbb{C}=\{0,1,\ldots,C\}$ is a set of class indices indicating whether a voxel is empty ($0$) or belongs to a specific class ($1,\ldots,C$).

As our approach is based on TTA, we assume the existence of a pre-trained SSC model $\mathcal{F}$ as follows:
\begin{equation}
    \mathbf{p} = \mathcal{F}(\mathbf{X}),
\end{equation}
where $\mathbf{p}\in[0,1]^{(C+1) \times L\times W \times H}$ is the probability of each voxel belonging to each class.
The final class prediction $\hat{\mathbf{Y}}$ can be obtained by applying an argmax function on $\mathbf{p}$.
In this context, the goal of TALoS is to adjust the pre-trained model to adapt to an arbitrary test sample $\mathbf{X}$ by optimizing the parameters, making them more suitable.
Here, note that the proposed approach does not have explicit requirements on $\mathcal{F}$, such as the architectures or pre-training policies.

\subsection{Test-time Adaptation via Line of Sight (TALoS)}\label{Sec:talos}

The proposed TALoS targets a realistic application of the pre-trained SSC model, assuming an autonomous vehicle drives through arbitrary environments in test time.
In this scenario, we suppose the point clouds captured by LiDAR are continuously provided as time proceeds, and accordingly, the model should perform SSC for each given point cloud instantly.
We denote the input sequence of point clouds as $\{\mathbf{X}_i\}$, where $i=\{1,\ldots\}$ indicates the moment when the point cloud is captured.

The main idea behind TALoS is that for guiding the model prediction of a certain moment (let $i$), the observation made at another moment (let $j$) can serve as supervision.
However, as the ego-vehicle moves as time proceeds, the input point clouds captured at the two different moments, \textit{i.e.}, $\mathbf{X}_i$ and $\mathbf{X}_j$, are on different LiDAR coordinates.
To handle this, we use a transformation matrix $\mathbf{T}_{j\rightarrow i}$ between two coordinates to transform the $j$th point cloud $\mathbf{X}_j$ with respect to the $i$th coordinate system, by
\begin{equation}\label{Equ:do_proj}
    \mathbf{X}_{j\rightarrow i} = \mathbf{T}_{j\rightarrow i} \mathbf{X}_j,
\end{equation}
where $\mathbf{X}_{j\rightarrow i}$ is the transformed point cloud.

\begin{figure}[t]
  \centering
  \includegraphics[width=0.99\linewidth]{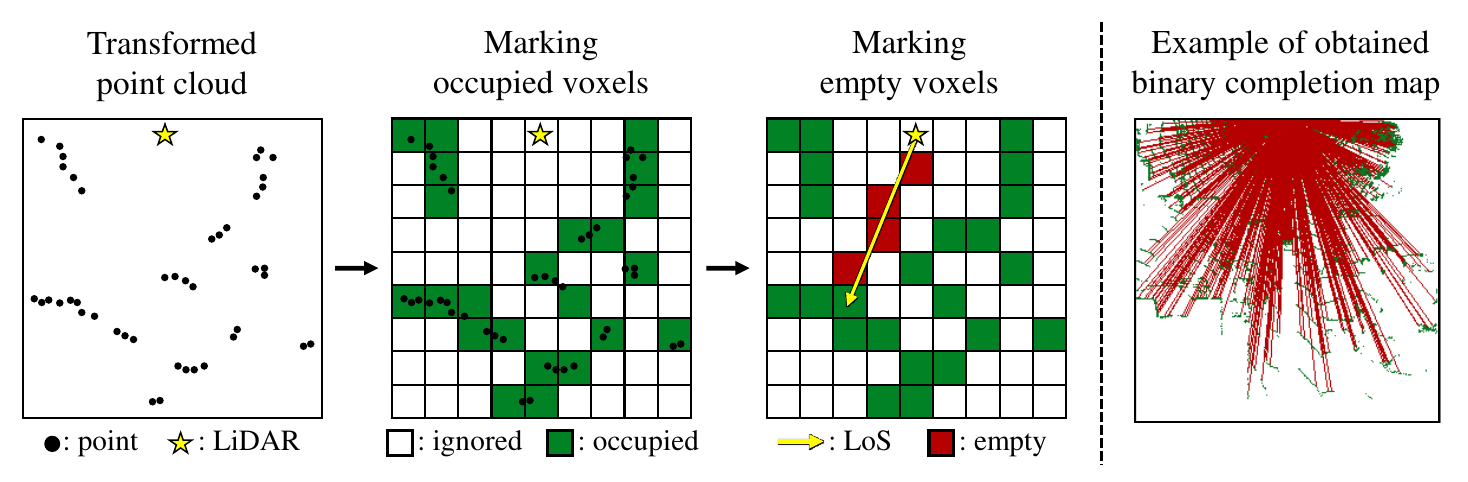}
  \vspace{-5pt}
  \caption{\textbf{Left}: Visualization of constructing a binary map $\mathbf{V}^{comp}_{j\rightarrow i}$ from the transformed point cloud $\mathbf{X}_{j\rightarrow i}$. Although we represent our process using a 2D grid for intuitive visualization, note that the real process is performed on a 3D voxel. \textbf{Right}: The real example of the binary map projected on 2D.}
  \label{Fig:los}
  \vspace{-10pt}
\end{figure}

Subsequently, we exploit $\mathbf{X}_{j\rightarrow i}$ to obtain a binary self-supervision for geometric completion, indicating whether a voxel is empty or occupied.
We implement this process as shown in Fig.~\ref{Fig:los}. 
For voxelization, we first define a voxel grid of a pre-defined size ($L\times W \times H$), initialized with an ignore index ($255$). 
Then, using $\mathbf{X}_{j\rightarrow i}$, we set the value of voxels containing at least one point to 1 (green color in Fig.~\ref{Fig:los}), while keeping the other voxels as $255$. 
Here, note that we discard the points of the non-static classes (\textit{e.g.}, car), as such object can change their location between $i$th and $j$th observation.
For this rejection, we use $\mathbf{p}_j$, the model prediction at $j$th moment.
The resulting binary mask indicates which voxels are occupied at the $j$th moment with respect to the $i$th coordinate.

Additionally, we use the Line of Sight (LoS), an idea utilized in various fields~\cite{hu2020you, ding2019deepmapping, chen2023deepmapping2}, to further identify which voxels should not be occupied. 
Considering LiDAR's characteristics, the space between the LiDAR and the voxels filled with 1 should be empty. 
To check which voxels are crossed by the LoS, we employ Bresenham's algorithm~\cite{bresenham1965algorithm}. 
For this process, we use the LiDAR position at the $j$th moment converted to the $i$th coordinate system, not the $i$th LiDAR position.
Finally, we set the value of the identified voxels to 0 (red color in Fig.~\ref{Fig:los}), indicating that the voxels should be empty. 

The obtained $\mathbf{V}^{comp}_{j\rightarrow i}\in\{0,1,255\}^{L\times W\times H}$ then serves as supervision for $\mathbf{p}_i$, the $i$th prediction.
Here, $\mathbf{p}_i$ is obtained by the pre-trained SSC model $\mathcal{F}$ as follows:
\begin{equation}
    \mathbf{p}_i = \mathcal{F}(\mathbf{X}_i).
\end{equation}
Since $\mathbf{p}_i$ is the prediction for all the classes, including the empty class, we convert it into the binary prediction for completion, denoted as $\mathbf{p}^{comp}_i\in[0,1]^{2\times L\times W\times H}$.
Here, the first element of $\mathbf{p}^{comp}_i$ is simply $\mathbf{p}^0_i$, and the second one is $\max_{c=1,\ldots,C}\mathbf{p}^c_i$, the maximum value among the scores of the non-empty classes.
Here, $\mathbf{p}^c_i$ represents the predicted probability of the voxels belonging to $c$th class at $i$th moment.

Finally, the loss function using the binary completion map is as
\begin{equation}\label{Equ:loss_comp}
    \mathcal{L}^{comp}_{j \rightarrow i}(\mathbf{p}_i) = \mathcal{L}_{ce}(\mathbf{p}^{comp}_i, \mathbf{V}^{comp}_{j\rightarrow i})+\mathcal{L}_{lovasz}(\mathbf{p}^{comp}_i, \mathbf{V}^{comp}_{j\rightarrow i}),
\end{equation}
where $\mathcal{L}_{ce}$ and $\mathcal{L}_{lovasz}$ are cross-entropy loss and lovasz-softmax loss~\cite{berman2018lovasz}, respectively.

\subsection{Extension for semantic perception}\label{Sec:semseg}

In the previous section, we described a method that leverages an observation made in one moment ($j$) to obtain the binary occupancy map of another moment ($i$).
As the method mainly focuses on enhancing the model's capability of understanding the test-time scene structure, \textit{i.e.}, scene completion, this section further extends our approach to address the semantic perception of surroundings, another key goal of SSC.
Specifically, we carefully identify the reliable regions from the prediction of the pre-trained model at each moment, and then build a consensus among these predictions across various moments.

\begin{figure}[t]
  \centering
  \includegraphics[width=0.99\linewidth]{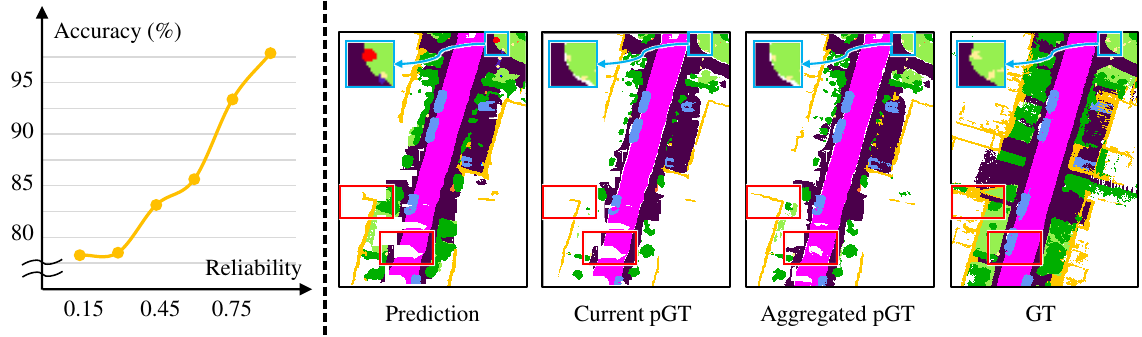}
  \caption{\textbf{Left}: Verification of the reliability metric. The voxels having higher reliability show higher semantic completion accuracy. \textbf{Right}: Examples of pseudo-GT (pGT) construction. The blue box depicts the successful rejection of misprediction using reliability, while the red boxes show the benefit of using the prediction of another moment, providing more completed pGT.}
  \label{Fig:reliability}
\end{figure}

To achieve this, we first define a metric similar to \cite{wang2020tent,liang2020we,niu2023towards}, based on Shannon entropy~\cite{shannon1948mathematical} $\mathcal{H}$, as follows:
\begin{equation}
    \mathbf{R} = 1 - {\mathbf{H}(p) \over \max_{p} \mathbf{H}(p)} \quad \text{where} \quad \mathcal{H}(\mathbf{p})=-\sum^{C}_{c=0} \mathbf{p}^c \log\mathbf{p}^c.
\end{equation}
Here, $\mathbf{R}\in[0,1]^{L\times W\times H}$ is the measured reliability.
As reported in conventional works based on confidence-based self-supervision, we also confirmed a high positive correlation between the reliability $\mathbf{R}$ and the actual accuracy of voxel-wise classification, as in Fig.~\ref{Fig:reliability} left.

To identify the confident regions from the model predictions, we simply threshold the measured reliability by a pre-defined value $\tau$ (0.75 in ours) as follows:
\begin{equation}\label{Equ:thresh}
    \mathbf{A}(x,y,z) =
  \begin{cases}
    \argmax_c \mathbf{p}^c(x,y,z) & \text{if} \hspace{3pt} \mathbf{R}(x,y,z)>\tau \\
    255 & \text{otherwise,} \\
  \end{cases} 
\end{equation}
where $(x,y,z)$ denotes the voxel coordinate.
Here, $\mathbf{A}\in (\mathbb{C}\cup\{255\})^{L\times W\times H}$ is the reliable self-supervision, which can function as pseudo-GT for semantic segmentation.

Following the above process, we first acquire $\mathbf{A}_j$ using $j$th model prediction.
Subsequently, we project its coordinate to $i$th coordinate, similar to Equ.~(\ref{Equ:do_proj}).
After obtaining the projected pseudo-GT, denoted as $\mathbf{A}_{j\rightarrow i}$, we aggregate it with $\mathbf{A}_i$, the pseudo-GT obtained at the current moment $i$.
In detail, we replace only the unconfident voxels in $\mathbf{A}_i$ (indexed as $255$) with the corresponding voxels of $\mathbf{A}_{j\rightarrow i}$.
Meanwhile, if the classes predicted by $\mathbf{A}_i$ and $\mathbf{A}_{j\rightarrow i}$ differ on certain voxels and both are confidently predicted, we conservatively drop those voxels as $255$.
As depicted in the colored boxes in Fig.~\ref{Fig:reliability} right, this aggregation helps our framework to build a consensus among the semantic perceptions performed at various moments, significantly enhancing the quality of pseudo-GT from the perspective of SSC.

We denote the result of aggregation as $\mathbf{V}^{sem}_{j\rightarrow i}\in (\mathbb{C}\cup\{255\})^{L\times W\times H}$.
Since $\mathbf{V}^{sem}_{j\rightarrow i}$ contains semantic information about all the classes, we can utilize it as direct pseudo-GT for guiding $\mathbf{p}_i$.
Accordingly, the loss function is defined as:
\begin{equation}\label{Equ:loss_sem}
    \mathcal{L}^{sem}_{j \rightarrow i}(\mathbf{p}_i) = \mathcal{L}_{ce}(\mathbf{p}_i, \mathbf{V}^{sem}_{j\rightarrow i})+\mathcal{L}_{lovasz}(\mathbf{p}_i, \mathbf{V}^{sem}_{j\rightarrow i}),
\end{equation}
where the notations are similar to those of Equ.~(\ref{Equ:loss_comp}).

\subsection{Dual optimization scheme for gradual adaptation}\label{Sec:dual}

The previous sections introduced how TALoS guides the prediction of a pre-trained SSC model at a certain moment ($i$), leveraging observations made at another moment ($j$). 
From a methodological perspective, the remaining step is to consider how to effectively adapt the model by selecting the appropriate moments.

Essentially, we cannot observe the future. 
Therefore, assuming TTA in real driving scenarios, we can only use past observations when updating the model at the current moment, which implies $i>j$. 
However, from the perspective of the SSC task, the main region of interest is intuitively the forward driving direction of the autonomous vehicle. 
This implies that guidance from future observations can be more important and valuable than guidance from past observations.

So, how can we leverage future information without actually observing it at the current moment?
Our key idea is to hold the model and its prediction at the current moment without updating and \textbf{delay the update until future observations become available}.

\begin{figure}[t]
  \centering
  \includegraphics[width=0.99\linewidth]{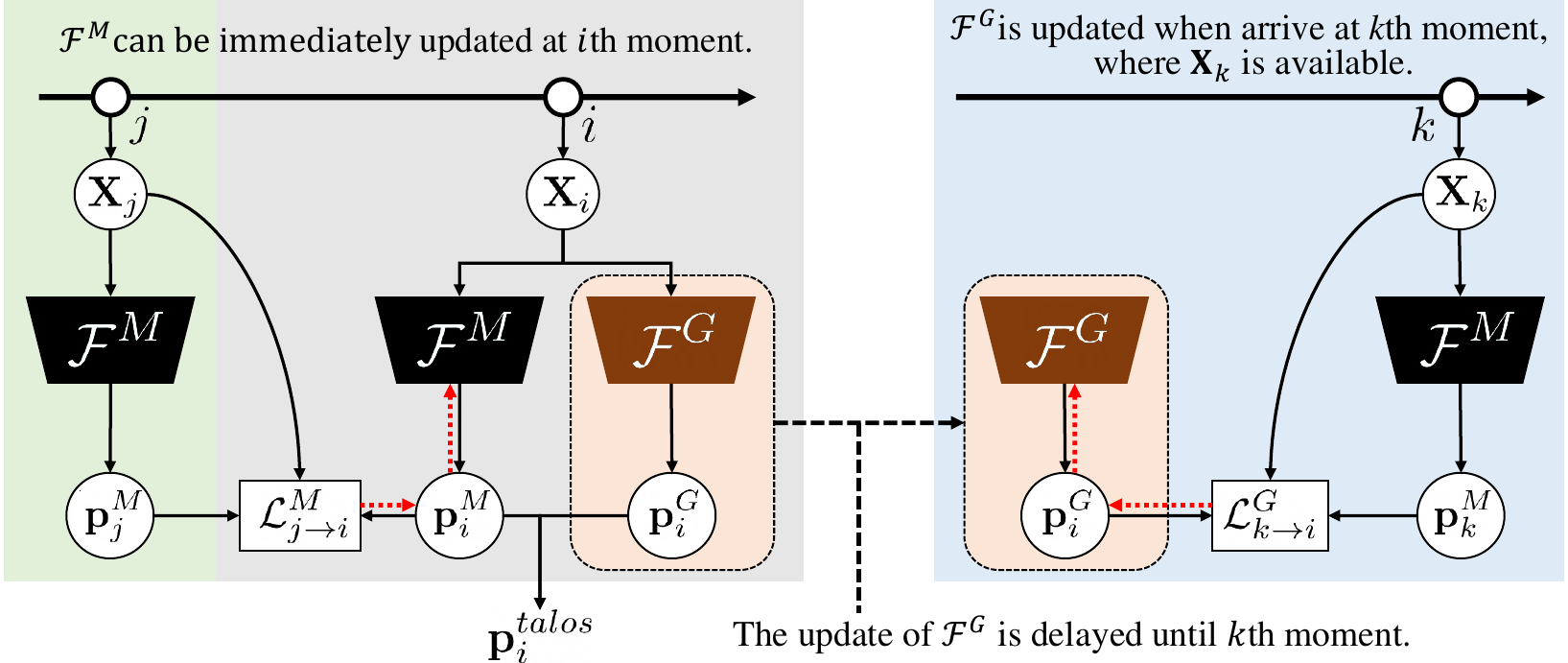}
  \caption{Conceptual visualization of the dual optimization scheme. $\mathcal{F}^M$ is instantly updated at moment $i$, using the past information provided from $j$th moment. On the other hand, the update of $\mathcal{F}^G$ using $i$th prediction is delayed until $k$th moment, when the future information becomes available. We unify the predictions of the models, $\mathbf{p}^M_i$ and $\mathbf{p}^G_i$, to get the final prediction $\mathbf{p}^{talos}_i$. The red dashed line denotes the back-propagation.}
  \label{Fig:dual}
  \vspace{-5pt}
\end{figure}

We develop this idea as in Fig.~\ref{Fig:dual}.
In detail, our approach involves two models of $\mathcal{F}^{M}$ and $\mathcal{F}^{G}$.
The goal of $\mathcal{F}^{M}$ is an instant adaptation on the sample of the current moment.
Therefore, we initialize $\mathcal{F}^{M}$ every moment with the pre-trained model and discard it at the end of the moment.
Here, $\mathcal{F}^{M}$ can be instantly updated at the moment $i$, using the past information already observed at $j$th moment.
The loss function for $\mathcal{F}^{M}$ is defined as:
\begin{equation}\label{Equ:loss_moment}
    \mathcal{L}^M_{j\rightarrow i}= \mathcal{L}^{comp}_{j \rightarrow i}(\mathbf{p}^M_i) + \mathcal{L}^{sem}_{j \rightarrow i}(\mathbf{p}^M_i),
\end{equation}
where $\mathbf{p}^M_i$ is the output of $\mathcal{F}^{M}$ at $i$th moment.

However, as aforementioned, we want to also leverage future information at $k$th moment, which is not available yet.
For this, we define $\mathcal{F}^{G}$, which aims to gradually learn the overall scene distribution.
Therefore, $\mathcal{F}^{G}$ is initialized only at the first step and continuously used for prediction. 
As in Fig.~\ref{Fig:dual}, the inference of $\mathcal{F}^{G}$ for $\mathbf{X}_i$ is instantly done and is used for the final output of the $i$th moment.
On the other hand, the update of $\mathcal{F}^{G}$ should stand by, until $\mathbf{X}_k$ is available.
In other words, once the model $\mathcal{F}^{G}$ arrives at $k$th moment (which was future at the time of prediction), the update is performed.
The loss function for $\mathcal{F}^{G}$ is defined as:
\begin{equation}\label{Equ:loss_cont}
    \mathcal{L}^G_{k\rightarrow i} = \mathcal{L}^{comp}_{k \rightarrow i}(\mathbf{p}^G_i) + \mathcal{L}^{sem}_{k \rightarrow i}(\mathbf{p}^G_i),
\end{equation}
where $\mathbf{p}^G_i$ is the output of $\mathcal{F}^{G}$ at $i$th moment.

This update cannot directly affect the prediction of $\mathcal{F}^{G}$ made at $i$th moment, as it is already over in the past from the perspective of $k$th moment when the update occurred.
However, this continuous accumulation of future information gradually enhances the model, allowing it to better learn the overall scene structure as time progresses. We provide a detailed illustration in Algorithm~\ref{alg:the_algotable}.

\begin{algorithm}[t]
\caption{Dual optimization scheme (single iteration)}
\begin{algorithmic}[1]
\STATE \textbf{Input:} Moment-model $\mathcal{F}^M$, Gradual-model $\mathcal{F}^G$, Pre-trained SSC model parameters $\theta_0$, temporal distance $\tau$, and Buffer $B$
\STATE \textbf{Output:} Adapted gradual-model parameters $\theta^G$
\STATE Initialize gradual-model parameters: $\theta^G \gets \theta_0$
\FOR{each timestep $t=1$ to unknown $T$}
    \STATE Initialize moment-model parameters: $\theta^M \gets \theta_0$
    \STATE $i,j \gets t,t-\tau$
    \STATE Receive current LiDAR observation $\mathbf{X}_i$
   
    \STATE Perform prediction with the moment-model: $\mathbf{p}^M_i \gets \mathcal{F}^M(\mathbf{X}_i;\theta^M)$
    \STATE Perform prediction with the gradual-model: $\mathbf{p}^G_i \gets \mathcal{F}^G(\mathbf{X}_i;\theta_i^G)$

    \STATE Save $\mathbf{p}^M_i$ in the buffer $B$. 
    \STATE Save $\mathbf{p}^G_i$ and the corresponding forward propagation graphs of $\mathcal{F}^G$ to the buffer $B$.

    \IF{$j>=1$}
    \STATE Load $\mathbf{p}^M_j$ from the buffer $B$. 
    \STATE Compute $\mathcal{L}^M_{j\rightarrow i}$ using $\mathbf{p}^M_i$ and $\mathbf{p}^M_j$
    \STATE Update moment-model parameters: $\theta^M \gets \theta^M - \eta \nabla_{\theta^M} \mathcal{L}^M_{j\rightarrow i}$

    \STATE Load $\mathbf{p}^G_j$ and the corresponding forward propagation graphs of $\mathcal{F}^G$ from the buffer $B$.
    \STATE Compute $\mathcal{L}^G_{i\rightarrow j}$ using $\mathbf{p}^G_i$ and $\mathbf{p}^G_j$
    \STATE Update moment-model parameters: $\theta^G \gets \theta^G - \eta \nabla_{\theta^G} \mathcal{L}^G_{i\rightarrow j}$
    
    \STATE Perform prediction with the updated moment-model: $\mathbf{p}^M_i \gets \mathcal{F}^M(\mathbf{X}_i;\theta^M)$
    \STATE Perform prediction with the updated gradual-model: $\mathbf{p}^G_i \gets \mathcal{F}^G(\mathbf{X}_i;\theta_i^G)$
    \ENDIF
    
    \STATE $\mathbf{p}_i^{talos} \gets \text{Agg}(\mathbf{p}^M_i, \mathbf{p}^G_i)$
    \STATE Return $\mathbf{p}_i^{talos}$ as the final SSC result of the current timestep $i$
\ENDFOR
\STATE Return adapted gradual-model parameters $\theta^G$
\end{algorithmic}\label{alg:the_algotable}
\end{algorithm}

In summary, the proposed TALoS framework involves two models, moment-wisely adapted $\mathcal{F}^M$ and gradually adapted $\mathcal{F}^G$.
To obtain the final prediction $\mathbf{p}^{talos}_i$ of $i$th moment, we individually run $\mathcal{F}^M$ and $\mathcal{F}^G$ using $\mathbf{X}_i$ as an input.
Both results $\mathbf{p}^M_i$ and $\mathbf{p}^G_i$ are unified into a single voxel prediction.
Here, we use $\mathbf{p}^M_i$ as a base, while trusting $\mathbf{p}^G_i$ only for the voxels predicted as static categories (such as roads or buildings) by $\mathbf{p}^G_i$.
The rationale behind this strategy is that continual adaptation makes $\mathcal{F}^{G}$ gradually adapt to the overall sequence, leading to a better understanding of the distribution of static objects.
We empirically found that the continuous adaptation is more facilitated for the static pattern than the movable objects having diverse distribution. 

\section{Experiments}\label{Sec:exp}



\subsection{Settings}
\paragraph{Datasets \& Metrics.}
We primarily experiment on SemanticKITTI~\cite{behley2019semantickitti}, the standard benchmark for SSC, comprising 22 LiDAR sequences.
Sequences 00 to 10 are used for pre-training SSC models, except for 08, which is employed as a validation set.
For testing, we use sequences 11 to 21.
Additionally, we verify TALoS on cross-dataset evaluation from SemanticKITTI to SemanticPOSS~\cite{pan2020semanticposs}.
From the 6 sequences of SemanticPOSS, we utilize only the validation sequence (02).
For more details, please refer to the supplementary material.
For evaluation, we employ intersection over union (IoU), a standard metric for semantic segmentation. We report both the completion IoU (cIoU) for binary occupancy prediction and the mean IoU (mIoU) for all classes.

\paragraph{Implementation.}
We employ the officially provided SCPNet~\cite{xia2023scpnet}, which is pre-trained on the SemanticKITTI~\cite{behley2019semantickitti} train set, as our baseline.
To apply TALoS in test time, we only update the last few layers of SCPNet.
Additionally, to prevent SCPNet's architecture from automatically making the voxels far from existing points empty, we use a 3D convolution layer to expand the region of sparse tensor computation. This ensures that distant voxels are properly involved in the test-time adaptation.
For optimization, we use Adam optimizers~\cite{kingma2014adam}, where the learning rates are set to 3e-4 and 3e-5 for $\mathcal{F}^{M}$ and $\mathcal{F}^{G}$, respectively.
For more details, refer to Section~\ref{Sec:supp_impl}.

\subsection{Ablation studies}
We conducted ablation studies to evaluate the effectiveness of each component of TALoS.
The configurations and results of the ablation studies are demonstrated in Table~\ref{Tab:abl}.
First, we verify the effectiveness of the loss functions we devised.
The result of Exp A confirms that minimizing $\mathcal{L}^{comp}$ indeed increases both cIoU and mIoU performance over the baseline, helping the model adapt to each test sample.
In addition, Exp B shows that using $\mathcal{L}^{sem}$ is also effective, especially for semantic perception, resulting in better mIoU performance.
Further, Exp C using both losses achieves even higher performance, demonstrating the effectiveness of the proposed TALoS.

Additionally, we check the validity of the dual optimization scheme, ablating either $\mathcal{F}^{M}$ or $\mathcal{F}^{G}$.
The results of Exp C and Exp D show that the use of moment-wise adaptation and gradual adaptation are both effective.
Further, Exp E confirms that our dual scheme effectively unifies both gains into our TALoS framework, significantly outperforming the baseline on both cIoU and mIoU.

\begin{table}[t]
\scriptsize
\centering
\caption{The results of ablation studies for the proposed TALoS framework, conducted on SemanticKITTI val set. COMP and SEM denote the use of loss function defined in Equ.~(\ref{Equ:loss_comp}) and Equ.~(\ref{Equ:loss_sem}), respectively. Meanwhile, MOMENT and GRADUAL represent the use of $\mathcal{F}^{M}$ and $\mathcal{F}^{G}$ for the optimization scheme in Sec.~\ref{Sec:dual}, respectively. All metrics are in \%. Best results are in \textbf{bold}.}
\label{Tab:abl}
\vspace{5pt}
\resizebox{0.99\textwidth}{!}{
\setlength{\tabcolsep}{12pt}
\begin{tabular}{c|cc|cc|c|c} 
\hline
\rowcolor{lightgray} & \multicolumn{2}{c|}{Loss functions} & \multicolumn{2}{c|}{Dual optimization scheme} & \multicolumn{2}{c}{Metrics} \\
\hline
\rowcolor{lightgray} & COMP & SEM & MOMENT & GRADUAL  &   mIoU  &  cIoU      \\ 
\hline
Baseline &  &    &    &     & 37.56 & 50.24 \\ \hline
A  & \checkmark &    &  \checkmark  &     & 37.97 & 52.81\\ \hline
B  & & \checkmark &  \checkmark  &    & 38.35 & 52.47 \\ \hline
C  & \checkmark & \checkmark &  \checkmark  &  & 38.38   & 52.95     \\ \hline
D  & \checkmark & \checkmark &  & \checkmark  & 38.81& 55.94 \\ \hline
E (Ours)  & \checkmark & \checkmark &  \checkmark  & \checkmark &  \textbf{39.29}&  \textbf{56.09} \\ \hline
\end{tabular}
}
\vspace{0pt}
\end{table}

\begin{figure}[t]
  \centering
  \vspace{-5pt}
  \includegraphics[width=0.99\linewidth]{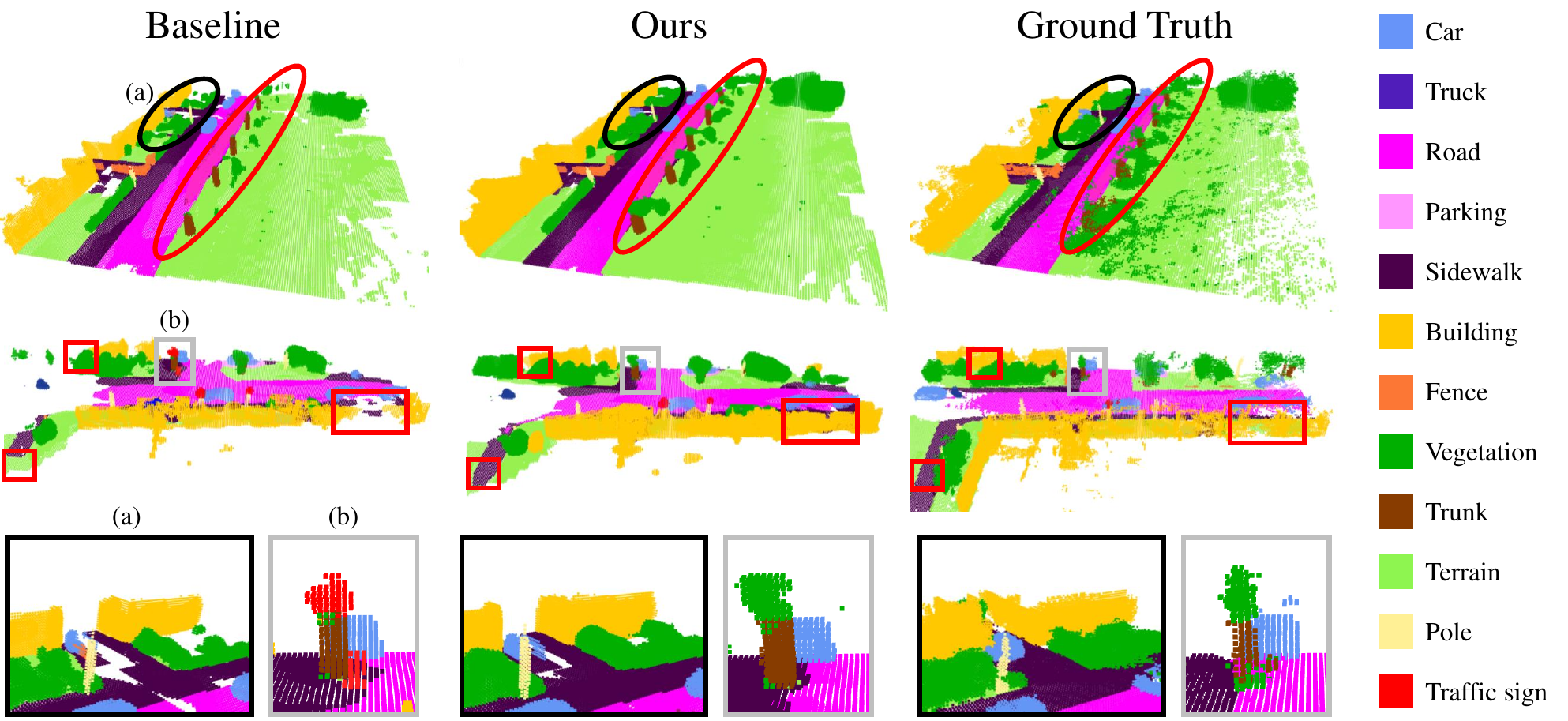}
  \vspace{0pt}
  \caption{Qualitative comparisons between baseline (SCPNet) and ours TALoS on SemanticKITTI val set. The highlighted regions depict the improvements achieved by TALoS, better completing the scene while also recovering the mispredictions.}
  \label{Fig:qual_kitti}
  \vspace{-5pt}
\end{figure}

\begin{table}[t]
\scriptsize
\centering
\caption{Quantitative comparison between the existing SSC methods with our TALoS on SemanticKITTI~\cite{behley2019semantickitti} test set. We use an online benchmark server for evaluation.}
\label{Tab:kitti}
\setlength\tabcolsep{2pt}
\resizebox{0.99\linewidth}{!}{
\begin{tabular}{l|c|c|ccccccccccccccccccc} 
\hline
       Methods   &               \rotatebox{90}{mIoU}                  &        \rotatebox{90}{cIoU}                         &          \rotatebox{90}{car}                 &                 \rotatebox{90}{bicycle}                   &         \rotatebox{90}{motorcycle}                            &                    \rotatebox{90}{truck}                   &                \rotatebox{90}{other-vehicle\hspace{4pt}}                    &             \rotatebox{90}{person}                    &           \rotatebox{90}{bicyclist}                            &      \rotatebox{90}{motorcyclist}                            &                     \rotatebox{90}{road}              &              \rotatebox{90}{parking}                    &                \rotatebox{90}{sidewalk}                       &                      \rotatebox{90}{other-ground}             &       \rotatebox{90}{building}   &       \rotatebox{90}{fence}  &       \rotatebox{90}{vegetation}   &       \rotatebox{90}{trunk}  &       \rotatebox{90}{terrain}  &       \rotatebox{90}{pole}  &       \rotatebox{90}{traffic-sign}                      \\ 
\hline\hline
SSA-SC~\cite{yang2021semantic} & 23.5 & 58.8 & 36.5 & 13.9 & 4.6 & 5.7 &   7.4 & 4.4 & 2.6 & 0.7 & 72.2 & 37.4 & 43.7 & 10.9 & 43.6 & 30.7 & 43.5& 25.6 & 41.8 & 14.5 & 6.9\\ 
JS3C-Net~\cite{yan2021sparse} & 23.8 & 56.6 & 33.3 & 14.4 & 8.8 & 7.2 &   12.7 & 8.0 & 5.1 & 0.4 & 64.7 & 34.9 & 39.9 & 14.1 & 39.4 & 30.4 & 43.1 & 19.6 & 40.5 & 18.9 & 15.9 \\ 
S3CNet~\cite{cheng2021s3cnet} & 29.5 & 45.6 & 31.2 & \textbf{41.5} & \textbf{45.0} & 6.7 &    16.1 & \textbf{45.9} & \textbf{35.8} & \textbf{16.0} & 42.0 & 17.0 & 22.5 & 7.9 & \textbf{52.2} & 31.3 & 39.5 & 34.0 & 21.2 & 31.0 & 24.3 \\ 
\hline
SCPNet~\cite{xia2023scpnet} & 36.7 & 56.1 & \textbf{46.4} & 33.2 & 34.9 & 13.8 &    29.1 & 28.2 & 24.7 & 1.8 & 68.5 &\textbf{51.3} & 49.8 & \textbf{30.7} & 38.8 & 44.7 & 46.4 & \textbf{40.1} & 48.7 & 40.4 & \textbf{25.1}\\ \hline
\rowcolor{lightgray} TALoS & \textbf{37.9}&\textbf{60.2}&\textbf{46.4}&34.4&36.9&\textbf{14.0}&\textbf{30.0}&30.5&27.3&2.2&\textbf{73.0}&\textbf{51.3}&\textbf{53.6}&28.4&40.8&\textbf{45.1}&\textbf{50.6}&38.8&\textbf{51.0}&\textbf{40.7}&24.4\\ \hline

\end{tabular}
}
\vspace{-10pt}
\end{table}

\begin{table}[t]
\centering
\setlength{\tabcolsep}{2pt}
\caption{Comparisons between the proposed TALoS and the existing TTA methods~\cite{wang2020tent,wang2022continual}. We use SCPNet baseline and conduct evaluation on SemanticKITTI val set.}
\label{Tab:tent}
\scriptsize
\resizebox{0.65\textwidth}{!}{
\begin{tabular}{l|c|cc|cc} 
\hline
\rowcolor{lightgray} Methods & Baseline & TENT~\cite{wang2020tent} & Ours (Exp. C) & CoTTA~\cite{wang2022continual} & Ours \\
\hline
mIoU & 37.56 & 37.92 & 38.38 & 36.55 & 39.29  \\
\hline
cIoU & 50.24 & 49.86 & 52.95 & 50.61 & 56.09 \\
\hline
\end{tabular}
}
\vspace{-5pt}
\end{table}

\subsection{Results on SemanticKITTI}

Table~\ref{Tab:kitti} provides the performance of existing SSC methods on the SemanticKITTI test set. 
Compared with SCPNet, which serves as our baseline, the proposed TALoS achieves significantly higher performance on both cIoU and mIoU. 
Considering that SCPNet also involves knowledge distillation using future frames during training, this performance gain confirms that our TTA-based method is more effective for leveraging future information.
Figure~\ref{Fig:qual_kitti} provides a qualitative comparison between the baseline and ours, demonstrating the advantages of TALoS.

Additionally, it is noteworthy to mention the difference between ours and OccFiner~\cite{shi2024occfiner}. 
OccFiner is designed to refine the results of existing SSC methods in an offline manner. It first generates predictions for a LiDAR sequence using an SSC method and then fuses these predictions post-driving to refine the results. In contrast, TALoS aims to perform test-time adaptation instantly in an online manner. We assume the sequential sensing of LiDAR data during driving, and TALoS gradually enhances the model as the test-time adaptation progresses.
As both methods have advantages in their respective practical settings, we simply mention it here, rather than comparing them in Table~\ref{Tab:kitti}.



\subsection{Comparisons with the existing TTA methods}

To demonstrate the benefit of our approach from the perspective of TTA, we integrated existing TTA studies into our framework and tested them. 
The results can be found in Table~\ref{Tab:tent}.
First, we performed optimization via entropy minimization, as in TENT~\cite{wang2020tent}, instead of minimizing the proposed loss functions. 
For this experiment, we exclusively use $\mathcal{F}^M$ for clear comparison.
This setting achieved 37.92\% and 49.86\% in mIoU and cIoU, respectively.
Note that cIoU of this setting is slightly lower than that of the baseline.
Further, the setting of Exp C in Table~\ref{Tab:abl}, which also uses $\mathcal{F}^M$ only, still outperforms TENT in both metrics.
These highlight the effectiveness of our losses, which utilize observations made at various moments.

To further clarify the superiority of our dual optimization scheme, we also implemented CoTTA~\cite{wang2022continual}, a continuous TTA approach based on the widely used student-teacher scheme.
We update both $\mathcal{F}^M$ and $\mathcal{F}^G$ are optimized using entropy minimization, where the update of $\mathcal{F}^G$ is assisted by CoTTA scheme.
We verify that this setting achieves 36.55\% of mIoU and 50.61\% of cIoU, where the mIoU decreases from the baseline. 
The results strongly confirm the superiority of the proposed optimization goals and schemes, which effectively leverage the information from driving scenarios for SSC.

\subsection{Experiments under the severe domain gap}
We check the potential of TALoS under the test scenarios of a target domain considerably different from the source domain. 
Specifically, we tested SCPNet pre-trained on SemanticKITTI, on the driving sequences of SemanticPOSS. 
Unlike SemanticKITTI, which uses a 64-beam LiDAR, SemanticPOSS is captured with a 40-beam and targets campus rather than on typical roads, resulting in significantly different class distribution.
Note that although TTA-based approaches could enhance performance by adapting the pre-trained model to the scene, the capability of the initial model itself is still essential.

Table~\ref{Tab:poss} compares our performance with that of the baseline, which uses the pre-trained SCPNet without any adjustments. 
Unfortunately, due to the severe drastic gap between SemanticKITTI and SemanticPOSS, the mIoU performance is actually low for both methods, as we expected.
Therefore, in this section, we would like to focus on the significant improvement achieved by TALoS over the baseline.
In particular, TALoS shows its potential by achieving an improvement of over 10 in cIoU, which is less affected by changes in class distribution.
We believe that combining TALoS with the prior studies targeting domain gaps could be an interesting direction for future research.

\subsection{Additional experiments}

\begin{figure}[H]
  \centering
  \vspace{-10pt}
  \includegraphics[width=0.99\linewidth]{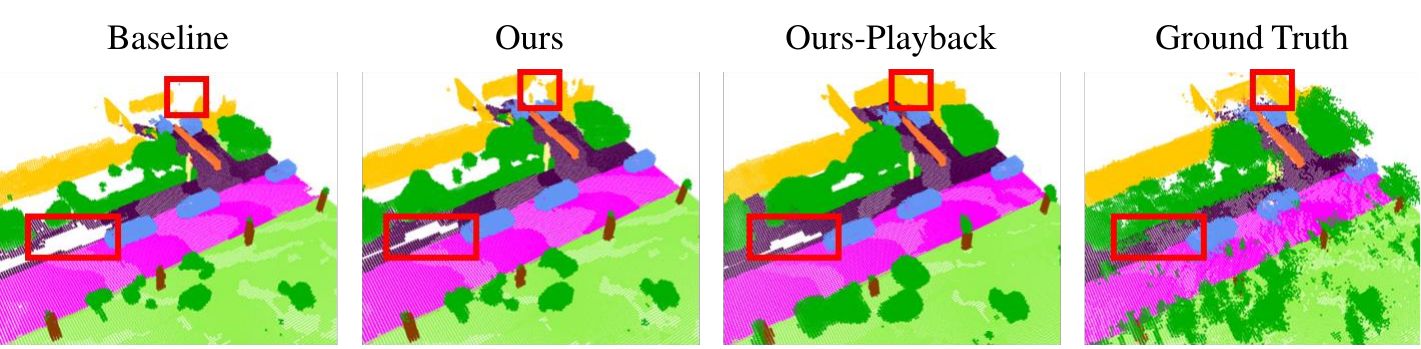}
  \vspace{-5pt}
  \caption{The results of the playback experiment. The red boxes depict the sequential improvements, implying that the gradual model indeed adapts to the scene as TTA proceeds.}
  \label{Fig:playback}
\end{figure}

\paragraph{Playback experiment.}
We further clarify the advantages of our continual approach with an intuitive experiment named ``playback''. 
Specifically, we first run TALoS on the SemanticKITTI validation sequence, from start to the end.
During this first round, we save the weights of the gradual model $\mathcal{F}^{G}$ at a certain moment. 
Subsequently, we initialize the $\mathcal{F}^{G}$ with the saved weights, and run TALoS once again on the same sequence from the start.
Here, during this playback round, we do not update the continual model at all.
If the $\mathcal{F}^{G}$ indeed learned the distribution of the scene while not being biased to a certain moment during the first round, the playback performance would be better than that of the first round.
Table~\ref{Tab:playback} verifies this expectation, where the playback performs better compared to not only the baseline but also the first round of TALoS.
We also provide a qualitative comparison between their results in Fig.~\ref{Fig:playback}, where the prediction of playback is clearly enhanced in terms of SSC.

\begin{table}[t]
\centering
\setlength{\tabcolsep}{2pt}
\caption{Results of the cross-dataset evaluation, pre-training on SemanticKITTI~\cite{behley2019semantickitti}, and evaluating on SemanticPOSS~\cite{pan2020semanticposs}. We compare the performance of TALoS with the baseline (SCPNet).}
\label{Tab:poss}
\vspace{5pt}
\scriptsize
\resizebox{0.75\textwidth}{!}{
\begin{tabular}{l|c|c|ccccccccccc} 
\hline
Methods   &               \rotatebox{90}{mIoU}                  &          \rotatebox{90}{cIoU}            &             \rotatebox{90}{person}        &      \rotatebox{90}{rider}                 &          \rotatebox{90}{car}           &       \rotatebox{90}{trunk}             &       \rotatebox{90}{plants}            &       \rotatebox{90}{traffic-sign\hspace{3pt}}            &       \rotatebox{90}{pole}              &       \rotatebox{90}{building}   &       \rotatebox{90}{fence}  &       \rotatebox{90}{bike}          &                     \rotatebox{90}{ground}                        \\ 
\hline\hline
Baseline (SCPNet~\cite{xia2023scpnet})   &  7.6  &  25.2  &  0.8  & 0.3   &  2.5  &  2.9  & 16.8  & 0.6  & \textbf{9.5} &  28.7  & \textbf{10.0}    & 3.8   & 7.2     \\ \hline
Ours   & \textbf{9.6} &  \textbf{36.2} &  \textbf{1.0}  &  \textbf{0.6}  &  \textbf{3.5}  &  \textbf{3.5}  &  \textbf{27.7}  &  \textbf{0.9}  &  8.2  &  \textbf{31.6}  &  6.1  &  \textbf{9.4}  &  \textbf{13.2}    \\ \hline
\end{tabular}
}
\end{table}

\begin{figure}[t]
\noindent
\begin{minipage}[t]{0.425\linewidth}
\centering
\captionof{table}{Results of the playback exp.}
\vspace{-6pt}
\resizebox{0.99\textwidth}{!}{
\setlength{\tabcolsep}{3pt}
\begin{tabular}{c|c|c|c} 
\hline
\rowcolor{lightgray} Methods & Baseline & Ours & Ours-Playback \\ 
\hline
cIoU & 50.24 & 56.09 & \textbf{56.91} \\
\hline
mIoU & 37.56 & 39.29 & \textbf{39.38} \\
\hline
\end{tabular}
\label{Tab:playback}
}
\end{minipage}
\begin{minipage}[t]{0.565\linewidth}
\centering
\captionof{table}{Impact of the number of iterations.}
\vspace{-6pt}
\resizebox{0.99\textwidth}{!}{
\setlength{\tabcolsep}{3pt}
\begin{tabular}{c|c|c|c|c|c} 
\hline
\rowcolor{lightgray} \# of iterations & 0 (baseline) & 1 & 2 & 3 (ours) & 5 \\ 
\hline
cIoU & 50.24 & 55.99 & 56.05 & \textbf{56.09} & 56.07 \\
\hline
mIoU & 37.56 & 39.09 & 39.22 & 39.29 & \textbf{39.31} \\
\hline
\end{tabular}
\label{Tab:iter}
}
\end{minipage}
\end{figure}

\paragraph{Impact of number of iterations.}
Table~\ref{Tab:iter} shows the impact of the number of iterations for updating $\mathcal{F}^M$ on the performance of TALoS.
Notably, TALoS achieves significant gains in both mIoU and cIoU, even with a single iteration.
Performance is enhanced as the number of iterations increases; however, we observe saturation after five iterations per sample.
The results imply that the proposed method efficiently excavates the information we targeted, fully leveraging it with only a few iterations.

\section{Conclusion}
This paper pioneers a novel Semantic Scene Completion (SSC) approach based on Test-time Adaptation (TTA).
The proposed method, named TALoS, focuses on that observation made at one moment can serve as Ground Truth (GT) for scene completion at another moment.
For this, we present several approaches to acquiring self-supervision that can be helpful in adapting the model to the scene in test time, from both the perspectives of geometric completion and semantic segmentation.
To further capitalize on future information that is inaccessible at the current time, we introduced a dual optimization scheme that delays the model update until future observations become available.
Evaluations on the Semantic-KITTI validation and test sets confirmed that TALoS significantly enhances the performance of the pre-trained SSC model.

One of our approach's main limitations is that it addresses point clouds only.
As there exist a number of SSC approaches using image data, exploring TTA for SSC in this setting could be an interesting research direction.
We believe that the main philosophy of this paper, using the observation of one moment to guide the prediction of another moment, can be seamlessly extended to the image-based approaches, enhancing the practicality of SSC.


\clearpage
\newpage
\appendix
\setcounter{table}{0}
\renewcommand\thetable{\thesection.\arabic{table}}
\renewcommand\thefigure{\thesection.\arabic{figure}}


\section{Implementation Details}\label{Sec:supp_impl}

This section provides more details about the implementation of our method. 
We utilize a voxel size of $256 \times 256 \times 32$, following SCPNet~\cite{xia2023scpnet}. 
For cross-domain evaluation, we use the class mapping from semanticKITTI~\cite{behley2019semantickitti} to semanticPOSS~\cite{pan2020semanticposs} shown in Table~\ref{tab:supp_map}, following~\cite{kim2023single,alonso2020domain}.  All experiments are conducted using a single NVIDIA RTX A6000.

Meanwhile, during the test-time adaptation using TALoS, we only updated the segmentation sub-network and fixed the weights of the other modules of SCPNet.
We experimentally observe that updating the whole network achieves slightly higher performance but is marginal.

\begin{table}[h]
    \centering
    \caption{Class mapping from semanticKITTI~\cite{behley2019semantickitti} to semanticPOSS~\cite{pan2020semanticposs}}
    \label{tab:supp_map}
    \resizebox{0.99\linewidth}{!}{
    \begin{tabular}{c|c|c|c|c|c|c|c|c|c|c|c}
    \hline
 \cellcolor{lightgray}   KITTI &  car & bicycle & person & bicyclist & road, sidewalk & building & fence & vegetation & trunk  & pole & traffic sign  \\ \hline
    \cellcolor{lightgray}    POSS   & car & bike & person & rider & ground & building & fence & plants & trunk  & pole & traffic-sign  \\ \hline
    \end{tabular}
    }
\end{table}

\section{Class-wise Results on SemanticKITTI Validation Set}

We provide class-wise results on the SemanticKITTI validation set in Table~\ref{Tab: sup_val}.
Notably, the proposed TALoS outperforms the baseline (SCPNet\cite{xia2023scpnet}) in most categories, in addition to the significant margins in overall metrics, mIoU and cIoU.
These results strongly confirm the superiority of our method.

\begin{table}[h]
\scriptsize
\centering
\caption{Quantitative results on SemanticKITTI validation set. }
\label{Tab: sup_val}
\setlength\tabcolsep{2pt}
\resizebox{0.99\linewidth}{!}{
\begin{tabular}{l|c|c|ccccccccccccccccccc} 
\hline
       Methods   &               \rotatebox{90}{mIoU}                  &        \rotatebox{90}{cIoU}                         &          \rotatebox{90}{car}                 &                 \rotatebox{90}{bicycle}                   &         \rotatebox{90}{motorcycle}                            &                    \rotatebox{90}{truck}                   &                \rotatebox{90}{other-vehicle\hspace{4pt}}                    &             \rotatebox{90}{person}                    &           \rotatebox{90}{bicyclist}                            &      \rotatebox{90}{motorcyclist}                            &                     \rotatebox{90}{road}              &              \rotatebox{90}{parking}                    &                \rotatebox{90}{sidewalk}                       &                      \rotatebox{90}{other-ground}             &       \rotatebox{90}{building}   &       \rotatebox{90}{fence}  &       \rotatebox{90}{vegetation}   &       \rotatebox{90}{trunk}  &       \rotatebox{90}{terrain}  &       \rotatebox{90}{pole}  &       \rotatebox{90}{traffic-sign}                      \\ 
\hline\hline
SCPNet~\cite{xia2023scpnet} & 37.6 & 50.2 & 51.2 & \textbf{25.8} & 37.7 & 57.6 &    43.8 & 22.9 & 18.8 & 4.2 & 70.8 & \textbf{61.5} & 53.0 & 15.3 & 33.6 & 32.2 & 38.9 & 33.9 & 52.9 & 39.8 & \textbf{19.8} \\ \hline
\rowcolor{lightgray} TALoS & \textbf{39.3}  &  \textbf{56.1}	 & \textbf{51.9}	 & \textbf{25.8} & 	\textbf{38.7} & 	\textbf{60.1}	 & \textbf{46.1} & 	\textbf{24.0} & 	\textbf{19.9} & 	\textbf{5.3} & 	\textbf{75.0} & 	61.3	 & \textbf{55.2}	 & \textbf{17.0}	 & \textbf{36.7}	 & \textbf{33.1} & 	\textbf{44.6} & 	\textbf{35.0} & 	\textbf{57.7} & 	\textbf{40.1}	 & 18.9\\ \hline

\end{tabular}
}
\end{table}

\section{Result on various baseline models}\label{Sec:baseline}
We conducted experiments using different architectures to prove that TALoS is a universally useful approach to various SSC models. As shown in Table~\ref{Tab:baseline}, TALoS meaningfully enhances SSC performance (mIoU) across different architectures and datasets. We utilized the baseline weights trained from the training set of the dataset on which validation was to be performed. The results imply that TALoS can be a promising solution for SSC in various settings.

\begin{table}[h]
\centering
\setlength{\tabcolsep}{2pt}
\caption{Comparisons between the proposed TALoS and TENT~\cite{wang2020tent} using various SSC models.}
\label{Tab:baseline}\scriptsize
\resizebox{0.65\textwidth}{!}{
\begin{tabular}{c|c|c|c|c}
\hline
\rowcolor{lightgray} 
SSC method & Dataset & Baseline & TENT~\cite{wang2020tent} & TALoS       \\ \hline
SSCNet~\cite{song2017semantic}              & KITTI-360        & 17.0              & 17.0 (+0.0)   & 17.4 (+0.4) \\ \hline
SSA-SC~\cite{yang2021semantic}              & SemanticKITTI    & 24.5              & 24.8 (+0.2)   & 25.3 (+0.8) \\ \hline
SCPNet~\cite{xia2023scpnet}              & SemanticKITTI    & 37.6              & 37.9 (+0.3)   & 39.3 (+1.7) \\ \hline
\end{tabular}}
\end{table}

\section{Additional Experimental Results}\label{Sec:supp_exp}

All the experiments in Section~\ref{Sec:supp_exp} are conducted on the SemanticKITTI validation set.

\subsection{Impact of threshold}
We verify the impact of $\tau$, the value thresholding the reliability for obtaining pseudo-GT, as in Equ.~(\ref{Equ:thresh}).
As shown in Table~\ref{Tab:supp_reliability}, the proposed method is quite robust to the change of $\tau$, and both mIoU and cIoU are saturated after a certain point (0.75).
Based on this result, we set the thresholding value to 0.75 by default in the main paper.

\subsection{Effectiveness of noise}
We verify the robustness of the proposed method against the errors that possibly exist in LiDAR calibration, as shown in Table~\ref{Tab:supp_noise}.
We model the error by disturbing the transformation matrix in Equ.~\ref{Equ:do_proj} with the noise sampled from the Gaussian distributions, which have standard deviation values listed in the first column of the table.
Specifically, we add the noise to the angles of rotation and translation vectors, to acquire the disturbed projection matrices between the LiDAR coordinate systems.
The results show that performance decreases as the level of noise increases, as expected.
Nevertheless, we observe that the proposed TALoS achieves substantial performance even with the noise, still meaningfully higher than the baseline. 
These results imply the robustness and practicality of our method.




\subsection{Impact of selecting moments}
As the proposed TALoS explicitly leverages the observations made at various moments during adaptation, it is essential to select proper moments.
To verify this, we conduct experiments by varying the chosen moments, as shown in Table~\ref{Tab:supp_moment}.
For example, frame difference 1 in Table~\ref{Tab:supp_moment} denotes selecting $j=i-1$ and $k=i+1$ as moments.
Here, we need to specially handle the boundary cases, \textit{e.g.}, at the start of the sequence ($j$<0) or the last of the sequence ($k$ is larger than the number of all the samples in the sequence).
For these cases, we simply do not use any losses relevant to those moments. 
Furthermore, for frame difference 0, we use $\mathbf{A}_i$, the pseudo-GT from the current moment, as the only self-supervision.

The results in Table~\ref{Tab:supp_moment} show that 
performance decreases if the selected moments are too far from the current step, as the distant observation may not overlap with the current observation or overly force the model to learn completely unpredictable regions, leading to bias.
Considering these, we set the frame difference to 1 by default in our setting.

\begin{figure}[h]
\noindent
\begin{minipage}[t]{0.34\linewidth}
\centering
\captionof{table}{Impact of changes in $\tau$, the thresholding for reliability.}
\vspace{-6pt}
\resizebox{0.99\textwidth}{!}{
\setlength{\tabcolsep}{10pt}
\begin{tabular}{c|c|c} 
\hline
\rowcolor{lightgray} 
Reliability & mIoU & cIoU \\ 
\hline
0.65 & 39.14 & 55.42 \\ 
\hline
0.7 & 39.24  & 55.8 \\ \hline
0.75 & \textbf{39.29} & \textbf{56.09} \\ \hline
0.8 & 39.28 & 56.05  \\ 
\hline
\end{tabular}
\label{Tab:supp_reliability}
}
\end{minipage}
\hspace{1pt}
\begin{minipage}[t]{0.3\linewidth}
\centering
\captionof{table}{Effectiveness of noises conducted on semantic- KITTI validation set.}
\vspace{-6pt}
\resizebox{0.99\textwidth}{!}{
\setlength{\tabcolsep}{10pt}
\begin{tabular}{c|c|c} 
\hline
\rowcolor{lightgray} 
Noise & mIoU & cIoU \\ 
\hline
0.05 & 38.58 & 54.09 \\ 
\hline
0.03 & 38.76 & 54.49 \\ 
\hline
0 & \textbf{39.29} & \textbf{56.09}  \\ 
\hline
\end{tabular}
\label{Tab:supp_noise}
}
\end{minipage}
\hspace{1pt}
\begin{minipage}[t]{0.33\linewidth}
\centering
\captionof{table}{Impact of selecting different moments in TALoS.}
\vspace{-6pt}
\resizebox{0.99\textwidth}{!}{
\setlength{\tabcolsep}{10pt}
\begin{tabular}{c|c|c} 
\hline
\rowcolor{lightgray} 
Frame Diff.& mIoU & cIoU \\ 
\hline
0  & 38.66 & 54.81 \\ 
\hline
1 & \textbf{39.29} & \textbf{56.09}  \\ 
\hline
2 & 39.21 & 55.94 \\ 
\hline
3 &39.14 & 55.92 \\
\hline
4 &39.09 & 55.91 \\	
	
\hline
\end{tabular}
\label{Tab:supp_moment}
}
\end{minipage}
\end{figure}

\subsection{Comparison with Fusion-based Approaches}

To analyze the effect of $F^M$, we compare our experiments with temporal fusion-based approaches.
Specifically, we devise naive temporal fusion methods for the previous and current timesteps in two different ways, named early and late fusion. 
In early fusion, we merge the raw point clouds of both the previous and current timesteps and use the fused point cloud as input for our baseline (the pre-trained SCPNet). 
On the other hand, in late fusion, we separately obtain the predictions of the baseline at each timestep and aggregate the results of different timesteps at the voxel level. 
Here, when the predicted classes differ, we trust the one with lower entropy (which means higher confidence).

Table~\ref{Tab:fusion} compares the performance of these fusion-based approaches with that of TALoS. 
For a fair comparison, we also replicate the performance of Exp. C in Table~\ref{Tab:abl} of the main paper to the 5th row. 
Note that this setting exclusively uses $F^M$. 
The results show that the cIoU gain of TALoS exceeds the naive temporal fusions, both early and late. 
Also, for mIoU, the fusion-based methods even harm the mIoU performance, while TALoS achieves significant gain. 
Finally, it is noteworthy that Exp. C, with $F^M$ only, still outperforms all the other fusion-based methods. 
This is strong evidence that $F^M$ indeed performs something more and better than the naive temporal fusion.

\subsection{Computational overhead}
We provide the required time per step of the baseline (SCPNet), conventional TTA methods, and TALoS in Table~\ref{Tab:time}. We use
SCPNet baseline and conduct evaluation on SemanticKITTI val set. 
Given the significant performance gains of TALoS, the result shows a reasonable trade-off of computational overhead and performance.

\begin{table}[H]
\centering
\setlength{\tabcolsep}{2pt}
\caption{Comparison with naive temporal fusion-based approaches.}
\vspace{3pt}
\label{Tab:fusion}\scriptsize
\resizebox{0.4\textwidth}{!}{
\begin{tabular}{c|c|c}
\hline
\rowcolor{lightgray} 
Method & mIoU (\%) & cIoU (\%)  \\ \hline
Baseline     & 37.56     & 50.24   \\ \hline
Early fusion    & 35.86   & 52.85    \\ \hline
Late fusion     & 37.11    & 52.49  \\ \hline
Exp. C (with $F^M$ only)   & 38.38      & 52.95   \\ \hline
TALoS     & \textbf{39.29}      & \textbf{56.09}    \\ \hline
\end{tabular}}
\end{table}

\begin{table}[t]
\centering
\setlength{\tabcolsep}{2pt}
\caption{Time and performance for proposed TALoS and existing TTA methods~\cite{wang2020tent,wang2022continual}.}
\vspace{3pt}
\label{Tab:time}\scriptsize
\resizebox{0.5\textwidth}{!}{
\begin{tabular}{c|c|c|c}
\hline
\rowcolor{lightgray} 
Method & Time per step (s) & overhead (\%) & mIoU (\%)  \\ \hline
Baseline     & 2.26   & -    & 37.56    \\ \hline
TENT~\cite{wang2020tent}     & 4.09  & +81    & 37.92    \\ \hline
CoTTA~\cite{wang2022continual}     & 6.14   & +171    & 36.55    \\ \hline
TALoS     & 6.65   & +194    & \textbf{39.29}    \\ \hline
\end{tabular}}
\end{table}

\clearpage



\end{document}